\documentclass{article}

\usepackage[preprint, nonatbib]{neurips_2023}

\usepackage[utf8]{inputenc} 
\usepackage[T1]{fontenc}    
\usepackage{hyperref}       
\usepackage{url}            
\usepackage{booktabs}       
\usepackage{amsfonts}       
\usepackage{nicefrac}       
\usepackage{microtype}      
\usepackage{xcolor}         
\usepackage{graphicx}
\usepackage{epstopdf}
\usepackage{subfigure}
\usepackage{caption}
\usepackage{tikz}
\usepackage{multirow}
\usetikzlibrary{shapes,arrows}

\title{Meta-DM: Applications of Diffusion Models on Few-Shot Learning}

\author{%
  Wentao Hu, Xiurong Jiang, Jiarun Liu, Yuqi Yang, Hui Tian\thanks{corresponding author} \\
  School of Information and Communication Engineering,\\
  Beijing University of Posts and Telecommunications\\
  \texttt{\{huwt, jiangxiurong, liujiarun01, yangyuqi, tianhui\}@bupt.edu.cn} \\
}

\begin{document}

\maketitle

\begin{abstract}
   In the field of few-shot learning (FSL), extensive research has focused on improving network structures and training strategies. However, the role of data processing modules has not been fully explored. Therefore, in this paper, we propose Meta-DM, a generalized data processing module for FSL problems based on diffusion models. Meta-DM is a simple yet effective module that can be easily integrated with existing FSL methods, leading to significant performance improvements in both supervised and unsupervised settings. We provide a theoretical analysis of Meta-DM and evaluate its performance on several algorithms. Our experiments show that combining Meta-DM with certain methods achieves state-of-the-art results. 
\end{abstract}

\section{Introduction}
Deep neural network-based machine learning methods have achieved remarkable success in various image recognition related areas. However, traditional deep learning techniques typically require enormous amounts of labeled data. This poses two main challenges. Firstly, it can be difficult to collect sufficient data in certain cases, such as identifying rare diseases or endangered animals. Secondly, labeling data is often laborious and time-consuming. In an attempt to address these challenges, FSL and unsupervised FSL have emerged as promising solutions.

Few-shot image recognition tasks require the machine to be trained on a small amount of images. Many sophisticated ideas have been proposed to address this challenge, including metric learning \cite{i1,i2,i4}, gradient-based meta-learning \cite{i9}, differentiable optimization layers \cite{i11}, hypernetworks \cite{i14}, neural optimizers \cite{i16}, transductive label propagation \cite{i6}, neural loss learning \cite{i15}, Bayesian neural priors \cite{i17}. Current methods can already achieve quite high accuracy rates on FSL tasks, even approaching the effectiveness of traditional deep learning \cite{i7,i19,i10}.

Currently, research on FSL problems primarily focuses on improving network structures and training strategies \cite{i1,i2,i4,i9,i11,i14,i16,i6,i15,i17}. Insufficient data is a core issue in FSL tasks, making data augmentation a potential solution. However, traditional image augmentation methods, such as cropping, rotation, and sharpening, do not effectively capture the intra-class differences of new classes with few labelled samples \cite{i34}. To address this challenge, researchers have explored more sophisticated data augmentation techniques, including AutoEncoders, RandAugment \cite{i20}, etc., and have achieved promising results \cite{i10,i22}. Another approach, proposed by Zhang et al., is Meta-GAN (Meta-Generative Adversarial Networks), which uses dummy samples generated by GAN to help the model learn clearer decision boundaries between different classes \cite{i3}.

In this work, we propose a novel approach to address the problem of insufficient data in FSL tasks.  We utilize diffusion models, a type of generator that has better generalizability and controllability than traditional GANs \cite{i21,i26,i27,i28}, to generate pseudo-data.  We show that by controlling the parameters of diffusion models, we can simultaneously achieve data augmentation and decision boundaries sharpening, which can function as a significant module in FSL models.  Our proposed method, referred to as Meta-DM (Meta-Diffusion Models), is easy to implement and can be combined with various existing FSL algorithms.  We provide theoretical analysis to explain why the use of Meta-DM yields better results and evaluate our method on several classical and advanced FSL algorithms with comparable performance on several benchmarks \cite{i1,i7,i8,i10}.  We are surprised to find that the use of Meta-DM provides significant improvements over almost all current FSL methods and even achieves state-of-the-art performance on several algorithms, without requiring changes to the model details or hyperparameters.

In summary, our contributions can be listed as follows:

\begin{itemize}

	\item We propose a novel data processing module called Meta-DM, which utilizes diffusion models and can be seamlessly integrated with various FSL models.

	\item By analyzing the theoretical principles of Meta-DM, we explain why it is effective for FSL.

	\item We conduct a series of experiments to evaluate the effectiveness and versatility of Meta-DM on various FSL algorithms. Our results demonstrate that Meta-DM can achieve state-of-the-art performance in both supervised and unsupervised settings.

\end{itemize}

\section{Related Works}

\subsection{Supervised Few-Shot Learning}
\label{Supervised Few-Shot Learning}

\paragraph{Problem Definition} The FSL problem is commonly defined as the $N$-way $K$-shot problem: given a base dataset $\mathcal{D}_{base} = \{(x_i , y_i)\}$, the goal is to train a model that can effectively solve the downstream few-shot task $\mathcal{T}$, which consists of a support set $\mathcal{S} = \{(x_s, y_s)\}^{N \times K}_{s=1}$ for adaptation and a query set $\mathcal{Q} = \{x_q\}^Q _{q=1}$ for prediction, where $y_s$ is the class label of image $x_s$.

\paragraph{Prototypical Networks} The Prototypical Network is considered as one of the most classical solutions for FSL problems \cite{i1}. The main idea of Prototypical Networks is to find a feature prototype for each class by computing the mean vector of its embedded support points:

\begin{equation}\label{eqn-1} 
	\mathbf{c}_k=\frac{1}{\left|S_k\right|} \sum_{\left(x_i, y_i\right) \in S_k} f_\phi\left(x_i\right)
\end{equation}

For a query point $x$, Prototypical Networks compute the Euclidean distance between $x$ and each feature prototype, and then produce a distribution over all classes by applying a softmax function to the distances:

\begin{equation}\label{eqn-2} 
	p_\phi(y=k \mid x)=\frac{\exp \left(-d\left(f_\phi(x), \mathbf{c}_k\right)\right)}{\sum_{k^{\prime}} \exp \left(-d\left(f_\phi(x), \mathbf{c}_{k^{\prime}}\right)\right)}
\end{equation}

The objective of training is to minimize the negative logarithm of the true class probability $k$: $J(\phi)=-\log p_\phi(y=k|x)$, thereby enhancing the classifier's recognition capability.

\paragraph{P>M>F} P>M>F stands for the pipeline consisting of Pre-training, Meta-learning and Fine-tuning \cite{i7}, each of which can be instantiated with different pre-training algorithm or backbone architecture. For example, Hu et al. uses DINO \cite{i24} for pre-training and Prototypical Networks with ViT \cite{i29} for meta-learning, resulting in fairly good performance on standard benchmarks. Additionally, P>M>F includes optional fine-tuning modules for cross-domain FSL tasks.

\subsection{Unsupervised Few-Shot Learning}

\paragraph{Problem Definition} The definition of unsupervised FSL is similar to that of supervised FSL, with the difference being that only unlabeled data $\mathcal{D}_{base} = \{x_i\}$ is available for training unsupervised FSL models.

\paragraph{Meta-GMVAE} Meta-GMVAE (Meta-Gaussian Mixture Variational Autoencoders) \cite{i8} is a pioneering unsupervised FSL algorithm based on VAE (Variational Autoencoders) \cite{i30} and set-level variational inference using self-attention \cite{i31}. The core idea of Meta-GMVAE is to use multimodal prior distributions, a mixture of Gaussians, considering that each modality corresponds to a class in unsupervised FSL tasks. It uses the EM (Expectation-Maximization) algorithm to optimize the parameters of Gaussian Mixture Model.

\paragraph{UniSiam} UniSiam is an advanced unsupervised FSL algorithm that achieves state-of-the-art performance on several benchmarks \cite{i10}. Unlike in supervised FSL, where the goal is to maximize the mutual information $I(Z; Y)$ between the representation $Z$ and the label $Y$, UniSiam aims to maximize the mutual information $I(Z; X)$ between the representation $Z$ and the data $X$. This approach enables the model to learn a meaningful representation of each class, while also avoiding overfitting to the base classes.

\subsection{Diffusion Models}

Recently, diffusion models have gained increasing popularity as a class of generative models due to their powerful generative capabilities \cite{i21}. The generative procedure of diffusion models can be roughly divided into two steps, as illustrated in Figure \cite{i26}:

\paragraph{Forward process (also referred to as diffusion process)} In this process, noise from a standard Gaussian distribution is continuously added to the input data. Given a sample $s_0$, noise is iteratively added to it to obtain $s_1, s_2, \ldots, s_{T-1}, s_T$. As the number of iterations tends to infinity, the resulting sequence of samples tends towards an isotropic Gaussian distribution.

\paragraph{Reverse process} In contrast to the forward process, the reverse process involves continuously filtering out noise and recovering data from the noisy data. As previously defined, we begin by sampling from $s_T$, and then gradually backtrack to $s_{T-1}, s_{T-2}, \cdots, s_0$.

The diffusion model is designed to learn the task of removing a small fraction of noise from $s_t$ to obtain $s_{t-1}$, where $t$ denotes any timestep during the process. To achieve this, the model employs a function $\epsilon_\theta (s_t,t)$ \cite{i36} that predicts the noise component of $s_t$. The training samples consist of the data $s_t$ at a given moment, the timestep $t$, and the noise $\epsilon$, all of which can be sampled during the forward process. The training objective is to minimize the mean square error loss between the predicted noise $\epsilon_\theta (s_t,t)$ and the actual noise $\epsilon$ \cite{i21}, as shown in the equation below:

\begin{equation}\label{eqn-3}
	\mathcal{L}_{diffusion} = \mathbb{E}_{s_t,t,\epsilon}||\epsilon_\theta (s_t,t) - \epsilon||^2
\end{equation}

\tikzstyle{start} = [rectangle,rounded corners, minimum width=1cm,minimum height=1cm,text centered, draw=black,fill=red!30]
\tikzstyle{process} = [rectangle,minimum width=3cm,minimum height=1.4cm,text centered,text width =2.2654cm,draw=black,fill=blue!20]
\tikzstyle{mid} = [rectangle,rounded corners, minimum width=1cm,minimum height=1cm,text centered, draw=black,fill=yellow!30]
\tikzstyle{stop} = [rectangle,rounded corners, minimum width=1cm,minimum height=1cm,text centered, draw=black,fill=green!30]
\tikzstyle{arrow} = [thick,->,>=stealth]
\begin{figure*}
	\centering
	\begin{tikzpicture}[node distance=2cm]
		\node (s0) [start] {$s_0$};
		\node (Diffusion) [process,right of= s0, xshift=1cm] {Diffusion \\ $p (s_t | s_{t-1})$};
		\node (st) [mid,right of=Diffusion, xshift=1cm] {$s_T$};
		\node (Denoising) [process,right of = st, xshift=1cm] {Denoising \\ $\epsilon_\theta (s_{t-1} | s_t)$};
		\node (s0^) [stop,right of= Denoising, xshift=1cm] {$\hat{s_0}$};
		
		\draw [arrow] (s0) -- (Diffusion);
		\draw [arrow] (Diffusion) -- (st);
		\draw [arrow] (st) -- (Denoising);
		\draw [arrow] (Denoising) -- (s0^);
	\end{tikzpicture}
	\caption{The diffusion model gradually adds noise to the original data, and then trains a deep learning model to gradually remove the noise. }
	\label{fig1}
\end{figure*}
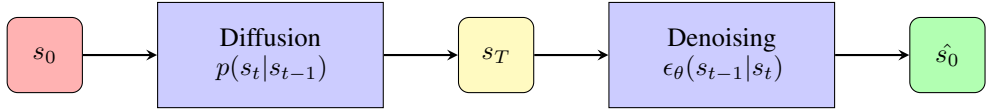

\section{Our Approach}

The central concept behind Meta-DM is to leverage diffusion models as generators of image-to-image transformations in order to obtain pseudo-data with varying degrees of similarity to the original data.  Specifically, high-similarity pseudo-data is utilized for data augmentation, while low-similarity pseudo-data is employed to sharpen decision boundaries. Note that we do not train a specific diffusion model tailored to a given algorithm or dataset.  Instead, we utilize the same diffusion model across all experiments, which ensures that Meta-DM can be readily applied to any new algorithm or dataset.  In the interest of maintaining a fair comparison, we keep all network architectures and hyperparameters constant throughout our experiments, except for data processing.

\subsection{Data Augmentation}

Our data augmentation module is simple to understand. Similar to traditional data augmentation methods, it treats the output of the generator as belonging to the same class as the original data. This approach helps the classifier to learn more information about each class. Specifically, we generate an augmented sample for each image and assign it to the original class. Our generator is denoted by $\mathcal{G}$, i.e., $\mathcal{G}(x_i, y_i) = (x_i', y_i)$.

\subsection{Decision Boundaries Sharpening}
\label{Decision Boundaries Sharpening}

Decision boundaries sharpening is an unconventional data processing technique that was proposed by Dai et al. \cite{i35}. Its effectiveness for FSL problems was demonstrated by Zhang et al. \cite{i3}. The technique allows the classifier not only to learn the features of each class, but also to better discriminate the feature boundaries between classes. For example, when classifying images of birds, the classifier needs to learn not only what a bird looks like, but also what images may resemble birds but are not birds, in order to reduce the probability of misclassifying an image such as an airplane, which shares some features with birds, as a "bird". 

\begin{figure*}
	\centering
	\subfigure[Original samples]{
		\begin{minipage}[t]{0.33\linewidth}
			\centering
			\includegraphics[width=1.2in]{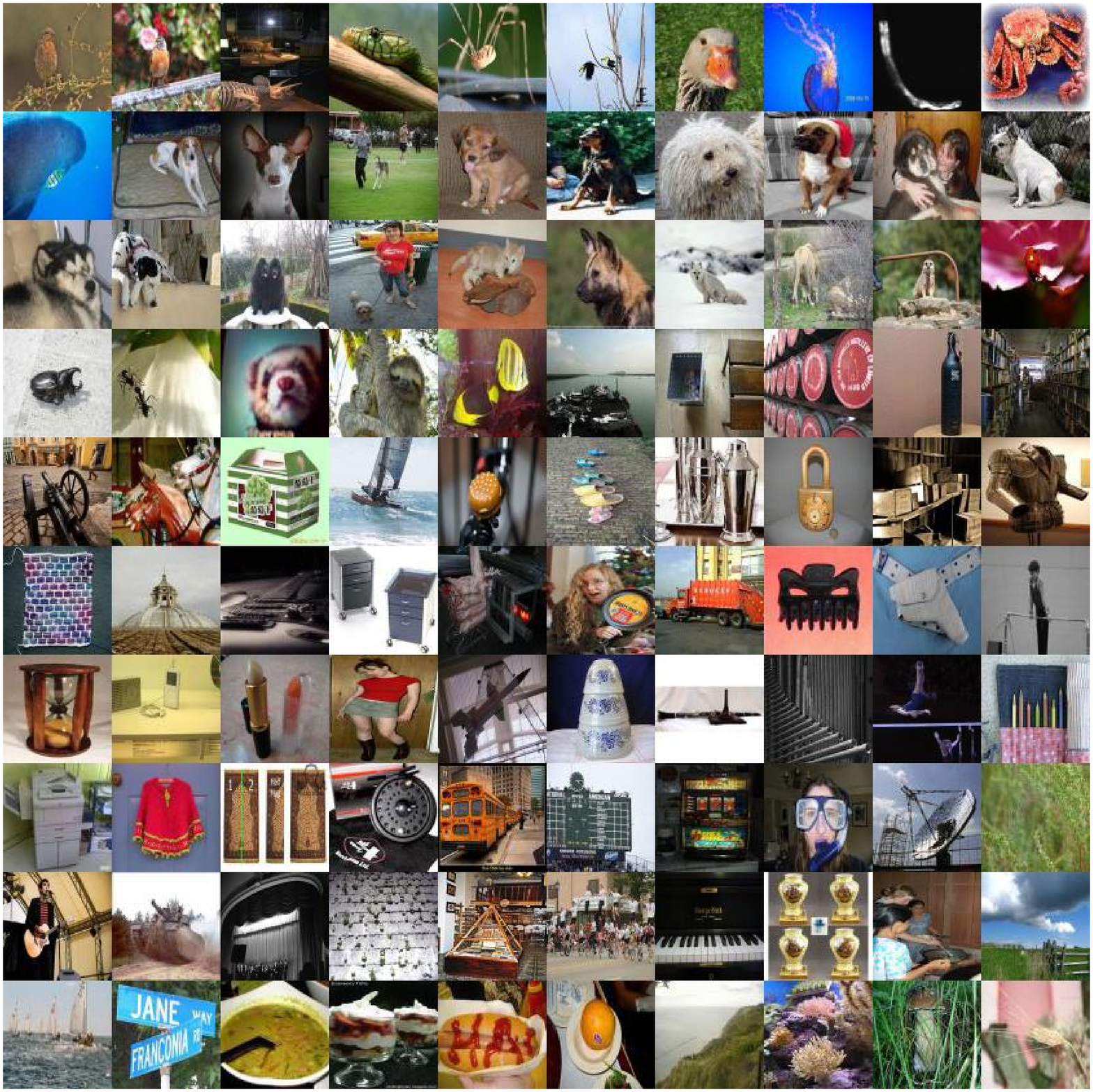}
		\end{minipage}
	}%
	\subfigure["Good" generated samples]{
		\begin{minipage}[t]{0.33\linewidth}
			\centering
			\includegraphics[width=1.2in]{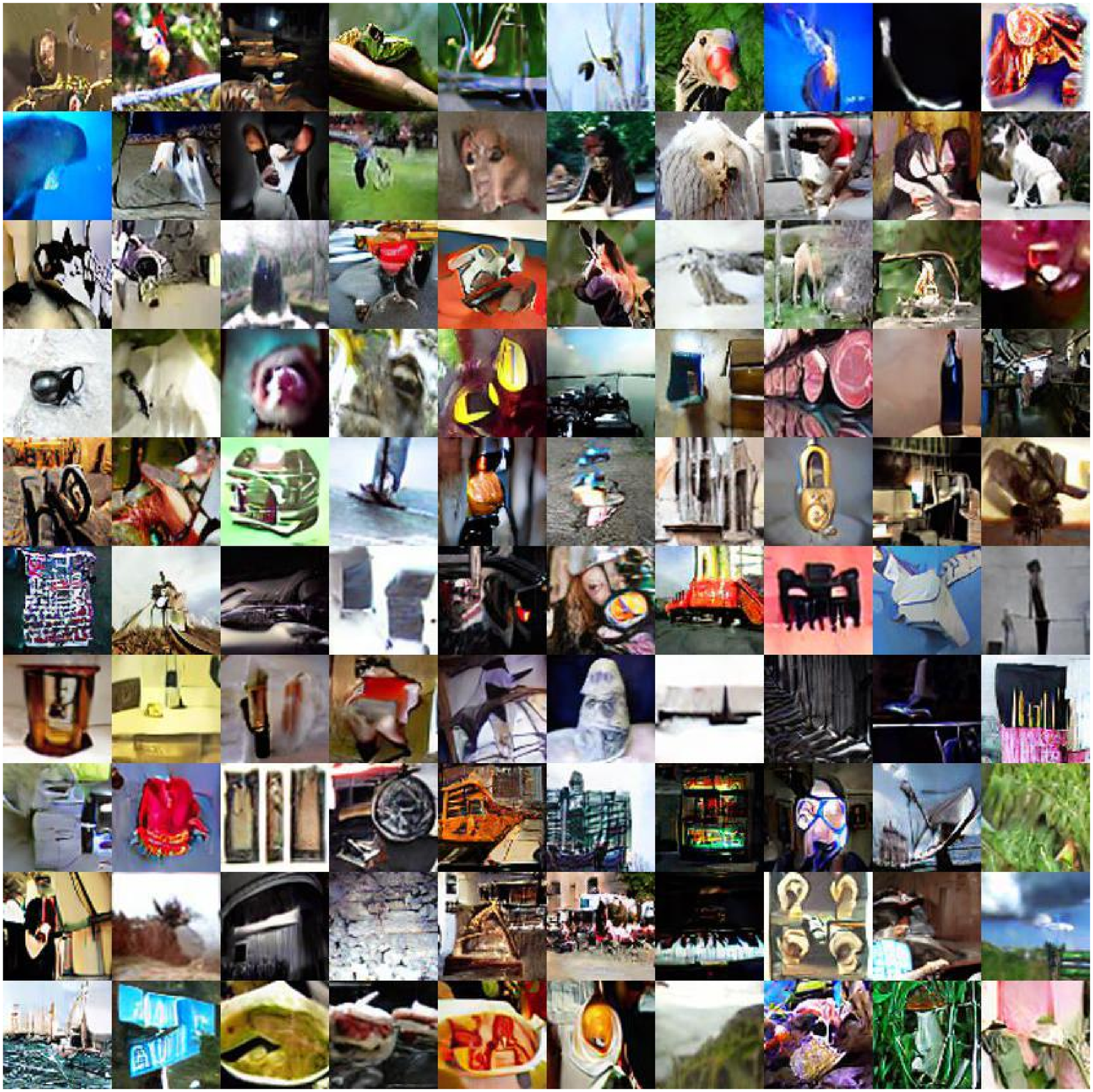}
		\end{minipage}
	}%
	\subfigure["Bad" generated samples]{
		\begin{minipage}[t]{0.33\linewidth}
			\centering
			\includegraphics[width=1.2in]{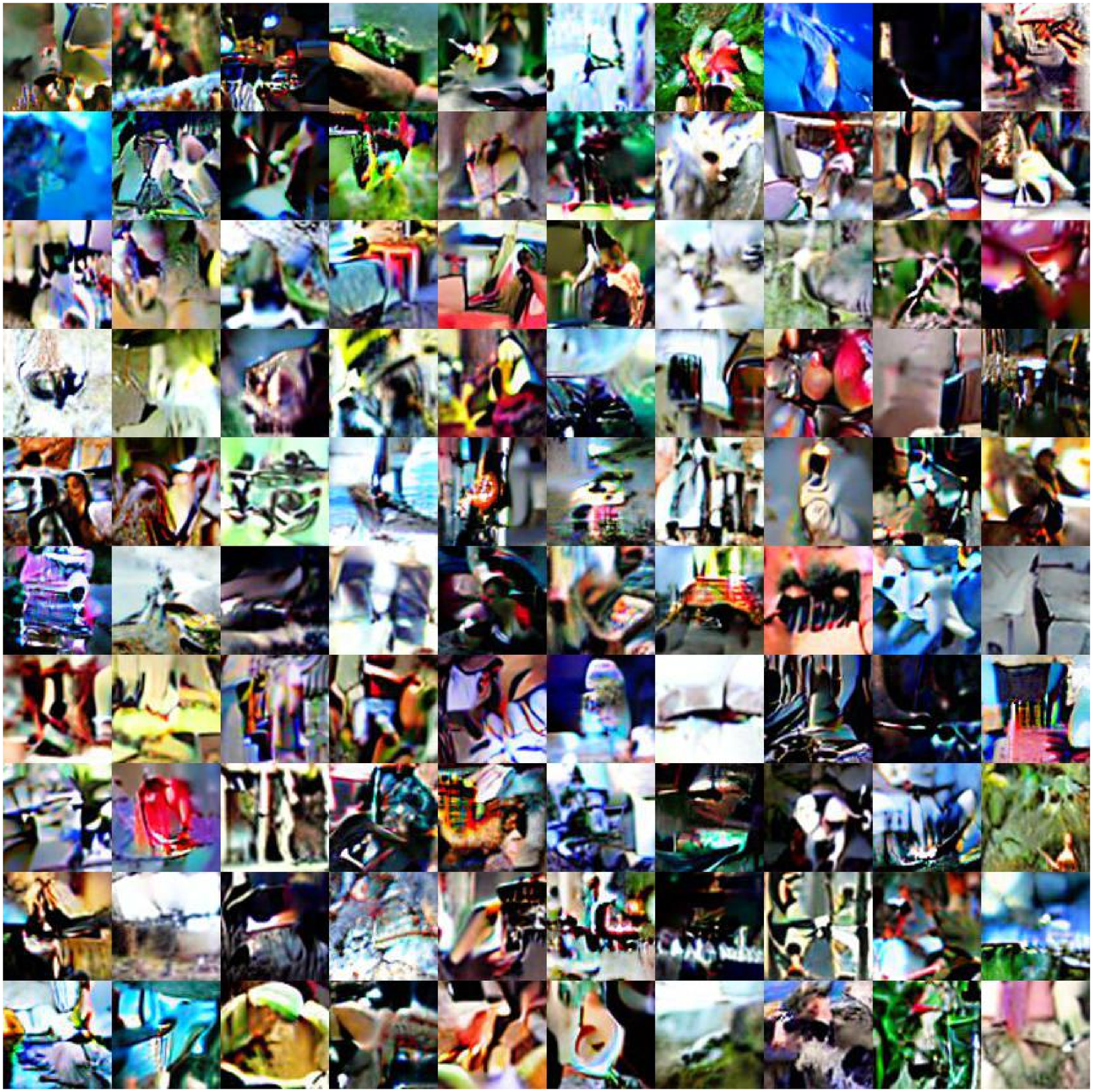}
		\end{minipage}
	}%
	\centering
	\caption{Comparison of the raw data and the pseudo-data generated by the two modules in Meta-DM. We obtain "good" and "bad" generated samples by adjusting the strength of diffusion models. Our generator is based on the design of von Platen et al. \cite{i25}. }
	\label{fig2}
\end{figure*}

To achieve this, "bad" generated samples that slightly deviate from the original data distribution are generated using the diffusion model, and are treated as additional classes that are different from the original classes, thus sharpening the decision boundaries from between original samples to between original samples and pseudo-samples. This is in contrast to the "good" generated samples used for data augmentation. We compare examples of "good" and "bad" samples in Fig \ref{fig2}.

Specifically, we offer two optional decision boundaries sharpening methods:

(1) Generate a "bad" sample based on each original image and put it into an "extra" class for each class, i.e., $\mathcal{G}(x_i, y_i) = (x_i', y_{fake-i})$.

(2) Randomly select a certain number of original images in each class for generation, and put all "bad" samples into the same "extra" class, i.e., $\mathcal{G}(x_i, y_i) = (x_i', y_{fake})$.

 During FSL training, the original data and the generated fake data can be combined to form a new training set. We can use a binary mask $m$ to distinguish between real and fake data. For example, $m=1$ for real data and $m=0$ for fake data. Then, the training loss can be defined as:

\begin{equation}
	\mathcal{L}_{FSL} = -\frac{1}{N}\sum_{i=1}^{N} m_i \log p(y_i|x_i) + \lambda ||\theta||^2
\end{equation}

where $N$ is the total number of samples, $x_i$ and $y_i$ are the input and output of the classifier, $p(y_i|x_i)$ is the predicted probability of the classifier, and $\lambda$ is a regularization parameter. For more proofs that decision boundaries sharpening is beneficial for image classification, please refer to Refs. \cite{i3,i35}.

\section{Experiments}

\subsection{Datasets}

\paragraph{\textit{mini}ImageNet} The \textit{mini}ImageNet dataset, originally proposed by Vinyals et al. \cite{i2}, is a subset of the widely used ImageNet dataset \cite{i37}. It consists of 60,000 color images in either their original size or in a processed size of 84$\times$84 pixels. The dataset is divided into 100 classes, with 600 images in each class. For our experiments, we follow the data splits proposed by Ravi and Larochelle \cite{i16}, where 64 classes are used for training, 16 classes for validation, and 20 classes for testing.

\paragraph{\textit{tiered}ImageNet} The \textit{tiered}ImageNet dataset, proposed by Ren et al. \cite{i6}, is a larger subset of ImageNet \cite{i37} than \textit{mini}ImageNet. It comprises 608 classes, each with about 1,300 images. These classes are further grouped into 34 higher-level categories, which are divided into 20 categories for training, 6 categories for validation, and 8 categories for testing.  Unlike other datasets, all classes in each category are used in only one stage, which ensures that all training classes are sufficiently fine-grained to differ from the test classes.

\paragraph{Meta-Dataset} The Meta-Dataset, proposed by Triantafillou et al. \cite{i44}, is a comprehensive benchmark dataset designed specifically for FSL. The dataset is constructed by combining images from 10 different datasets, including Omniglot, Aircraft, CUB-200-2011, etc. The Meta-Dataset is much larger than any previous FSL dataset, and each component image may have different types, styles, and sizes, which places higher demands on the generalizability of the FSL model \cite{i44}.

\subsection{Supervised FSL with Meta-DM}

\paragraph{Meta-DM $+$ Prototypical Networks} To allow the classifier to learn more features, we employ the full Meta-DM module on Prototypical Networks, which have a relatively simple backbone and do not use external data for pre-training. Specifically, for each original image, we generate a "good" sample and a "bad" sample, and assign the "good" sample to the original class and the "bad" sample to the "extra" classes. To generate the "good" samples, we set the strength of the diffusion model to 0.05, while for the "bad" samples, we set it to 0.2. Our approach is compared against Prototypical Networks without Meta-DM and several other classical supervised FSL algorithms, with the results presented in Table \ref{table1}. For more training details, please see our supplemental material.

\begin{table}
  \caption{Performance of Meta-DM $+$ Prototypical Networks and comparison to classical supervised FSL algorithms. All accuracy results are reported with 95$\%$ confidence intervals. }
  \centering
  \begin{tabular}{lccc}
    \toprule
      & \multicolumn{2}{c}{5-way Acc.} &  \\
    Method     & 1-shot     & 5-shot & dataset \\
    \midrule
    Matching Networks \cite{i2} & 43.40 $\pm$ 0.78   & 51.09 $\pm$ 0.71  & \multirow{5}{*}{\textit{mini}ImageNet}    \\
    MAML \cite{i9}     & 48.70 $\pm$ 1.84  & 63.15 $\pm$ 0.91  &     \\
    Relation Net \cite{i4} & 50.44 $\pm$ 0.82   & 65.32 $\pm$ 0.70  &  \\
    \cmidrule(r){1-3}
    Prototypical Nets \cite{i1} & 49.42 $\pm$ 0.78  & 68.20 $\pm$ 0.66  &  \\
    \textbf{Meta-DM $+$ Prototypical Nets} & \textbf{59.30 $\pm$ 0.29}  &  \textbf{ 72.28 $\pm$ 0.25 } &  \\
    \midrule
    Prototypical Nets \cite{i6} & 46.52 $\pm$ 0.52  & 66.15 $\pm$ 0.66  & \multirow{2}{*}{\textit{tiered}ImageNet} \\
    \textbf{Meta-DM $+$ Prototypical Nets} & \textbf{47.92 $\pm$ 0.45}  &  \textbf{ 69.09 $\pm$ 0.37 } &  \\
    \bottomrule
  \end{tabular}
  \label{table1}
\end{table}

\paragraph{Meta-DM $+$ P>M>F} As discussed in Section \ref{Supervised Few-Shot Learning}, there are multiple pre-training algorithms that can be used to instantiate P>M>F, such as DINO \cite{i24}, BEiT \cite{i47}, and CLIP \cite{i48}. We follow Hu et al. and use DINO for self-supervised pre-training. P>M>F also offers a variety of advanced backbone architectures, including ResNet-50 \cite{i46} and Vit \cite{i29}, which have stronger feature extraction capabilities. Given that the classifier is expected to have learned sufficient intra-class features during pre-training, we use a simplified version of Meta-DM for P>M>F. Specifically, we randomly select 5 images out of the 600 in each class and generate only "bad" samples based on them. All "bad" samples are assigned to an "extra" class, and the strength for "bad" samples is still 0.2. Since we train and test on the same dataset, fine-tuning is not necessary. Table \ref{table2} presents the results of Meta-DM $+$ P>M>F with ResNet-50 and Vit-Small/Vit-Base as backbones. In the above two experiments, Meta-DM demonstrate its ability to achieve broad improvements in supervised FSL problems. More importantly, it is able to provide considerable enhancements to state-of-the-art algorithms.

\begin{table}
	\caption{Performance of Meta-DM $+$ P>M>F on \textit{mini}ImageNet and comparison to representative state-of-the-art methods. }
	\centering
	\begin{tabular}{llccc}
		\toprule
		 & & \multicolumn{2}{c}{5-way Acc.} &  \\
		Method  & Backbone   & 1-shot     & 5-shot & Dataset \\
		\midrule
		Prototypical Nets \cite{i1} & Conv-4 & 49.42 $\pm$ 0.78  & 68.20 $\pm$ 0.66  &  \multirow{8}{*}{\textit{mini}ImageNet}    \\
		Meta-Baseline \cite{i38} & ResNet-12     & 63.17 $\pm$ 0.23 & 79.26 $\pm$ 0.17  &     \\
		PT-MAP \cite{i39} & WRN-28-10 & 82.92 $\pm$ 0.26    & 88.82 $\pm$ 0.13  &  \\
		CNAPS + FETI \cite{i40} & ResNet-18 & 79.9 $\pm$ 0.8  & 91.5 $\pm$ 0.4  &  \\
		\cmidrule(r){1-4}
		P>M>F \cite{i7} & ResNet-50 & 79.2 $\pm$ 0.4  & 92.0 $\pm$ 0.2  &  \\
		\textbf{Meta-DM $+$ P>M>F} & ResNet-50 & \textbf{81.05 $\pm$ 0.42}  &  \textbf{ 92.73 $\pm$ 0.19 } &  \\
		P>M>F \cite{i7} & Vit-Small & 93.1 $\pm$ 0.2  & 98.0 $\pm$ 0.1  &  \\
		\textbf{Meta-DM $+$ P>M>F} & Vit-Small & \textbf{93.82 $\pm$ 0.25}  &  \textbf{ 98.24 $\pm$ 0.08 } &  \\
		P>M>F  & Vit-Base & 95.3 $\pm$ 0.2  & 98.4 $\pm$ 0.1  &  \\
		\textbf{Meta-DM $+$ P>M>F} & Vit-Base & \textbf{95.65 $\pm$ 0.21}  &  \textbf{ 98.66 $\pm$ 0.07 } &  \\
		\bottomrule
	\end{tabular}
	\label{table2}
\end{table}

\begin{table}
	\caption{Comparison of cross-domain task results with and without Meta-DM on Vit-Small}
	\centering
	\begin{tabular}{lcccc}
		\toprule
		& \multicolumn{4}{c}{Dataset} \\
		& Birds     & Omniglot & Traffic Signs & Aircraft\\
		\midrule
		P>M>F \cite{i7}  & 86.38 $\pm$ 0.69  & 77.32 $\pm$ 1.41  & 92.53 $\pm$ 0.58 & 86.77 $\pm$ 0.57 \\
		\textbf{Meta-DM $+$ P>M>F}  & \textbf{86.48 $\pm$ 0.69}  &  \textbf{ 77.68 $\pm$ 1.38 } & \textbf{ 92.58 $\pm$ 0.57 } & \textbf{ 86.92 $\pm$ 0.58 } \\
		\bottomrule
	\end{tabular}
	\label{table3}
\end{table}

To gain the performance of Meta-DM on cross-domain tasks, we train the model on \textit{mini}ImageNet and then test it on several subsets of the Meta-Dataset. We follow Hu et al. by first automatically selecting the learning rate during the model deployment phase and then fine-tuning the feature backbone through several gradient steps \cite{i7}. Meta-DM is only used in the training stage. We compare the results of cross-domain tasks with and without Meta-DM in Table \ref{table3}. The results show that Meta-DM can also achieve a slight improvement on cross-domain tasks.

\subsection{Unsupervised FSL with Meta-DM}

\paragraph{Meta-DM $+$ Meta-GMVAE} In this experiment, we utilize the complete Meta-DM module that is used on Prototypical Networks. To ensure accurate comparison, we only report the performance results on \textit{mini}ImageNet, which includes 5-way, 1-shot / 5-shot / 20-shot / 50-shot settings. The performance results are shown in Table \ref{table4}.

\begin{table}
	\caption{Performance of Meta-DM $+$ Meta-GMVAE on \textit{mini}ImageNet and comparison to classical unsupervised FSL algorithms. }
	\centering
	\begin{tabular}{lcccc}
		\toprule
		& \multicolumn{4}{c}{5-way Acc.} \\
		Method  & 1-shot   & 5-shot & 20-shot & 50-shot \\
		\midrule
		UMTRA \cite{i41} & 39.93 $\pm$ 0.75  & 50.73 $\pm$ 0.71  & 61.11 $\pm$ 0.69 & 67.15 $\pm$ 0.62\\
		CACTUs-MAML \cite{i45}  & 39.90 $\pm$ 0.74  & 53.97 $\pm$ 0.70  & 63.84 $\pm$ 0.70 & 69.64 $\pm$ 0.63 \\
		CACTUs-ProtoNets  & 39.18 $\pm$ 0.71  & 53.36 $\pm$ 0.70  & 61.54 $\pm$ 0.68 & 63.55 $\pm$ 0.64 \\
		\midrule
		Meta-GMVAE \cite{i8}  & 42.82 $\pm$ 0.74  & 55.73 $\pm$ 0.64  & 63.14 $\pm$ 0.53 & 68.26 $\pm$ 0.49 \\
		\textbf{Meta-DM $+$ GMVAE}  & \textbf{52.09 $\pm$ 0.79}  &  \textbf{ 61.78 $\pm$ 0.69 } & \textbf{67.63 $\pm$ 0.62 }& \textbf{ 70.39 $\pm$ 0.60 } \\
		\bottomrule
	\end{tabular}
	\label{table4}
\end{table}

\paragraph{Meta-DM $+$ UniSiam} In the original UniSiam by Lu et al., the 224 $\times$ 224 image input size and two effective data augmentation methods: RandomVerticalFlip and RandAugment \cite{i20}, are used in order to improve the model's ability to extract features from scratch \cite{i10}. Therefore, we utilize the simplified Meta-DM module as in P>M>F, with the only difference being that the strength of the diffusion model is adjusted to 0.3 as the image input size increases. For accurate comparison, we report the performance of Meta-DM $+$ UniSiam with ResNet-18 / ResNet-34 / ResNet-50 as backbones and \textit{mini}ImageNet / \textit{tiered}ImageNet as benchmarks. To further push the limits of unsupervised FSL problems, we also conduct standard knowledge distillation \cite{i49} on each of our models, with the teacher models being the previous models we trained with Meta-DM $+$ UniSiam and ResNet-50. The results are shown in Table \ref{table5}. Our experiments demonstrate that Meta-DM also achieves state-of-the-art performance on unsupervised FSL. Specifically, it exhibits considerable improvements on \textit{mini}ImageNet and slight improvements on \textit{tiered}ImageNet. This suggests that Meta-DM is better suited for datasets with fewer classes and fewer samples per class, which facilitates the learning of new features by the classifier and ultimately leads to better performance. Furthermore, the increase in computational power requirement resulting from the generating a small number of "bad" samples is negligible compared to the original experiment.

\begin{table}
	\caption{Performance of Meta-DM $+$ UniSiam and comparsion to previous unsupervised FSL algorithms. The columns with "dist" in parentheses are the results with standard knowledge distillation \cite{i49}. }
	\centering
	\begin{tabular}{llccc}
		\toprule
		& & \multicolumn{2}{c}{5-way Acc.} &  \\
		Method  & Backbone   & 1-shot     & 5-shot & Dataset \\
		\midrule
		UMTRA \cite{i41} & ResNet-18 & 39.93 $\pm$ 0.75  & 50.73 $\pm$ 0.71  &  \multirow{15}{*}{\textit{mini}ImageNet}    \\
		ProtoCLR \cite{i42} & ResNet-18 & 50.90 $\pm$ 0.36 & 71.59 $\pm$ 0.29  &     \\
		UniSiam \cite{i10} & ResNet-18 & 63.26 $\pm$ 0.36  & 81.13 $\pm$ 0.26  &  \\
		\textbf{Meta-DM $+$ UniSiam} & ResNet-18 & \textbf{65.22 $\pm$ 0.37}  &  \textbf{ 82.18 $\pm$ 0.27 } &  \\
		UniSiam (\textit{dist}) & ResNet-18 & 64.10 $\pm$ 0.36  & 82.26 $\pm$ 0.25  &  \\
		\textbf{Meta-DM $+$ UniSiam (\textit{dist})} & ResNet-18 & \textbf{65.64 $\pm$ 0.36}  &  \textbf{ 83.97 $\pm$ 0.25 } &  \\
		\cmidrule(r){1-4}
		SimCLR \cite{i32} & ResNet-34 & 63.98 $\pm$ 0.37   & 79.80 $\pm$ 0.28  &  \\
		SimSiam \cite{i43} & ResNet-34 & 63.77 $\pm$ 0.38  & 80.44 $\pm$ 0.28  &  \\
		UniSiam \cite{i10} & ResNet-34 & 64.77 $\pm$ 0.37  & 81.75 $\pm$ 0.26  &  \\			
		\textbf{Meta-DM $+$ UniSiam} & ResNet-34 & \textbf{67.29 $\pm$ 0.39}  &  \textbf{ 83.30 $\pm$ 0.26 } &  \\
		UniSiam (\textit{dist})  & ResNet-34 & 65.55 $\pm$ 0.36  & 83.40 $\pm$ 0.24  &  \\		
		\textbf{Meta-DM $+$ UniSiam (\textit{dist})} & ResNet-34 & \textbf{66.22 $\pm$ 0.37}  &  \textbf{ 84.58 $\pm$ 0.25 } &  \\
		\cmidrule(r){1-4}
		UniSiam \cite{i10} & ResNet-50 & 65.33 $\pm$ 0.36  & 83.22 $\pm$ 0.24  &  \\
		\textbf{Meta-DM $+$ UniSiam} & ResNet-50 & \textbf{66.65 $\pm$ 0.36}  &  \textbf{ 84.26 $\pm$ 0.25 } &  \\		
		\textbf{Meta-DM $+$ UniSiam (\textit{dist})} & ResNet-50 & \textbf{66.68 $\pm$ 0.36}  &  \textbf{ 85.29 $\pm$ 0.23 } &  \\
		\midrule
		SimCLR \cite{i32} & ResNet-18 & 63.38 $\pm$ 0.42   & 79.17 $\pm$ 0.34  & \multirow{14}{*}{\textit{tiered}ImageNet} \\
		SimSiam \cite{i43} & ResNet-18 & 64.05 $\pm$ 0.40  & 81.40 $\pm$ 0.30  &  \\
		UniSiam \cite{i10} & ResNet-18 & 65.18 $\pm$ 0.39  & 82.28 $\pm$ 0.29  &  \\
		\textbf{Meta-DM $+$ UniSiam} & ResNet-18 & \textbf{65.31 $\pm$ 0.40}  &  \textbf{ 82.62 $\pm$ 0.30 } &  \\
		UniSiam (\textit{dist}) & ResNet-18 & 67.01 $\pm$ 0.39  & 84.47 $\pm$ 0.28  &  \\
		\textbf{Meta-DM $+$ UniSiam (\textit{dist})} & ResNet-18 & \textbf{67.11 $\pm$ 0.40}  &  84.39 $\pm$ 0.28 &  \\
		\cmidrule(r){1-4}
		UniSiam \cite{i10} & ResNet-34 & 67.57 $\pm$ 0.39  & 84.12 $\pm$ 0.28  &  \\			
		\textbf{Meta-DM $+$ UniSiam} & ResNet-34 & \textbf{67.74 $\pm$ 0.40}  &  \textbf{ 84.29 $\pm$ 0.29 } &  \\
		UniSiam (\textit{dist})  & ResNet-34 & 68.65 $\pm$ 0.39  & 85.82 $\pm$ 0.27  &  \\		
		\textbf{Meta-DM $+$ UniSiam (\textit{dist})} & ResNet-34 & \textbf{69.03 $\pm$ 0.41}  &  \textbf{ 85.90 $\pm$ 0.28 } &  \\
		\cmidrule(r){1-4}
		UniSiam \cite{i10} & ResNet-50 & 69.11 $\pm$ 0.38  & 85.82 $\pm$ 0.27  &  \\
		\textbf{Meta-DM $+$ UniSiam} & ResNet-50 & \textbf{69.53 $\pm$ 0.39}  &  \textbf{ 85.99 $\pm$ 0.27 } &  \\
		UniSiam (\textit{dist})  & ResNet-50 & 69.60 $\pm$ 0.38  & 86.51 $\pm$ 0.26  &  \\		
		\textbf{Meta-DM $+$ UniSiam (\textit{dist})} & ResNet-50 & \textbf{69.61 $\pm$ 0.38}  &  \textbf{ 86.53 $\pm$ 0.26 } &  \\
		\bottomrule
	\end{tabular}
	\label{table5}
\end{table}

\subsection{Ablation Studies}

To demonstrate the effectiveness of each module and the reasonableness of the parameters in Meta-DM, we conduct several ablation experiments. Specifically, we examine the effects of the completeness of Meta-DM, the generator, the diffusion model strength, and the number of "bad" samples.

\subsubsection{The Impact of the Completeness of Meta-DM}

To measure the impact of the completeness of Meta-DM, we conduct an ablation study of our Meta-DM-based data augmentation module and decision boundaries sharpening module on Prototypical Networks and \textit{mini}ImageNet, as shown in Table \ref{table6}. We find that when either module is used alone, there is an improvement over the initial algorithm, but neither is as effective as when both modules are used together. This suggests that both modules in Meta-DM have a positive effect. However, Table \ref{table6} also illustrates that the decision boundaries sharpening module leads to greater improvement. This is likely because, limited by the number of original images, it is difficult for the classifier to learn too many new features even with data augmentation, without changing the network structure. Therefore, we discard the data augmentation module in Meta-DM to reduce the computational cost on algorithms like P>M>F and UniSiam, which use more complex networks to extract features.

\begin{table}
	\caption{Ablations of different modules in Meta-DM on Prototypical Networks and \textit{mini}ImageNet. }
	\centering
	\begin{tabular}{lcccc}
		\toprule
		& & & \multicolumn{2}{c}{5-way Acc.}  \\
		Method  & "good" samples   & "bad" samples & 1-shot& 5-shot \\
		\midrule
		\multirow{4}{*}{Prototypical Nets \cite{i1}}  & & & 49.42 $\pm$ 0.78  & 68.20 $\pm$ 0.66  \\
		 & \checkmark & &  51.07 $\pm$ 0.63  & 69.41 $\pm$ 0.55  \\
		 &  & \checkmark &  59.20 $\pm$ 0.39  & 72.14 $\pm$ 0.37  \\
		 & \checkmark & \checkmark &  59.30 $\pm$ 0.47  & 72.28 $\pm$ 0.43  \\
		\bottomrule
	\end{tabular}
	\label{table6}
\end{table}

\subsubsection{The Impact of the Generator Type}

To evaluate the impact of the generator type in the decision boundaries sharpening module, we compare our Meta-DM-based approach with the GAN-based method proposed by Zhang et al. \cite{i3}. We apply both Meta-DM and Meta-GAN to two different FSL algorithms, Relation Networks and Soft k-Means, and compare their performance on \textit{mini}ImageNet in Table \ref{table7}. Our experiments demonstrate that Meta-DM outperforms Meta-GAN under the same experimental conditions. Additionally, Meta-GAN requires training a new generator for each algorithm or dataset by optimizing the discriminator's loss, while Meta-DM uses the same generator for various algorithms and datasets, which simplifies the application process. Furthermore, GANs are known to be difficult to train, and our initial work has shown that it can be challenging to train a suitable generator using Meta-GAN on sophisticated algorithms like UniSiam. In contrast, our Meta-DM approach remains effective on these state-of-the-art algorithms. In summary, our work has several advantages over Meta-GAN, including superior performance, easier combination with new algorithms and datasets, and robustness on challenging tasks.

\begin{table}
	\caption{Comparison of Meta-DM and Meta-GAN. }
	\centering
	\begin{tabular}{lccc}
		\toprule
		& & \multicolumn{2}{c}{5-way Acc.} \\
		Method &    & 1-shot     & 5-shot  \\
		\midrule
		Relation Networks \cite{i4} & \textit{sup.}  & 50.44 $\pm$ 0.82   & 65.32 $\pm$ 0.70    \\
		Meta-GAN $+$ Relation Net \cite{i3} & \textit{sup.}  & 52.71 $\pm$ 0.64   & 68.63 $\pm$ 0.67    \\
		Meta-DM $+$ Relation Net  & \textit{sup.}  & \textbf{56.97 $\pm$ 0.46}  & \textbf{69.47 $\pm$ 0.48}     \\
		\midrule
		Soft k-Means \cite{i6} & \textit{semi-sup.} & 50.09 $\pm$ 0.45   & 64.59 $\pm$ 0.28    \\
		Meta-GAN $+$ Soft k-Means \cite{i3} & \textit{semi-sup.} & 53.21 $\pm$ 0.89   & 66.80 $\pm$ 0.78    \\
		Meta-DM $+$ Soft k-Means & \textit{semi-sup.} & \textbf{61.91 $\pm$ 0.91}   & \textbf{72.13 $\pm$ 0.82}   \\
		\bottomrule
	\end{tabular}
	\label{table7}
\end{table}

\subsubsection{The Impact of the Diffusion Model Strgenth}

\begin{figure*}
	\centering
	\subfigure[The strength for data augmentation]{
		\begin{minipage}[t]{0.5\linewidth}
			\centering
			\includegraphics[width=2.5in]{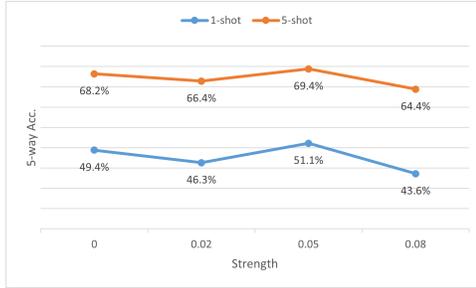}
		\end{minipage}
	}%
	\subfigure[The strength for decision boundaries sharpening]{
		\begin{minipage}[t]{0.5\linewidth}
			\centering
			\includegraphics[width=2.5in]{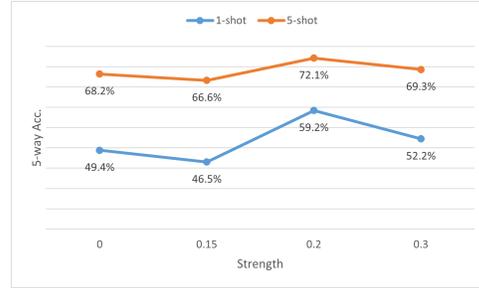}
		\end{minipage}
	}%
	\centering
	\caption{Ablations of the diffusion model strength on Prototypical Networks and \textit{mini}ImageNet. }
	\label{fig3}
\end{figure*}

The strength of the diffusion model determines the similarity between the generated samples and the original images. As the primary aim of Meta-DM is to generate samples that can aid in the classifier's training, setting the strength appropriately becomes critical. To determine the optimal value for the strength parameter, we conduct a series of ablation experiments on Prototypical Networks and \textit{mini}ImageNet, as depicted in Figure \ref{fig3}. Our experiments allow us to select the most effective strength value that maximizes the benefits of the generated samples in improving classifier performance.

\subsubsection{The Impact of the Number of "Bad" Samples}

\begin{figure*}
	\centering
	\subfigure[On Prototypical Networks]{
		\begin{minipage}[t]{0.5\linewidth}
			\centering
			\includegraphics[width=2.5in]{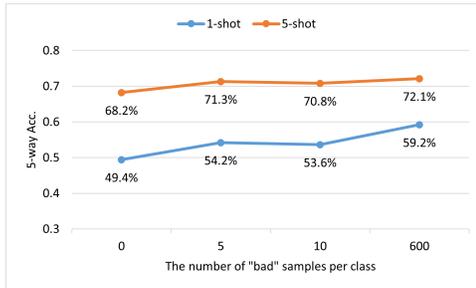}
		\end{minipage}
	}%
	\subfigure[On UniSiam with ResNet-50]{
		\begin{minipage}[t]{0.5\linewidth}
			\centering
			\includegraphics[width=2.5in]{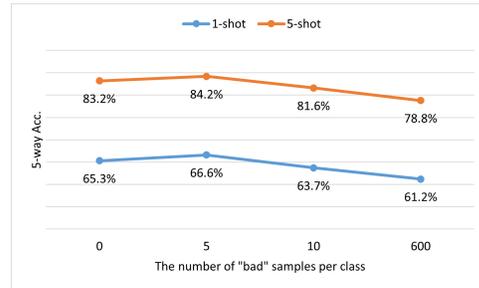}
		\end{minipage}
	}%
	\centering
	\caption{Ablations of the number of "bad" samples on \textit{mini}ImageNet. }
	\label{fig4}
\end{figure*}

As previously noted in Section \ref{Decision Boundaries Sharpening}, the choice of sharpening strategy can impact the number of "bad" generated samples. To further investigate this issue, we conduct ablation studies on the number of "bad" samples using Prototypical Networks and UniSiam as examples, as shown in Figure \ref{fig4}. Our results suggest that it is necessary to select different sharpening strategies for different algorithms to achieve optimal performance.

\section{Conclusions}

In this paper, we propose a novel data processing module for FSL, named Meta-DM, which uses diffusion models as image generators. Meta-DM can be easily incorporated into almost any FSL algorithm without changing any model details or hyperparameters, and has only a minor increase in computational cost. In addition to theoretical analysis, we experimentally demonstrate the effectiveness of Meta-DM by applying it to various existing supervised and unsupervised FSL algorithms and benchmarks. Our results demonstrate that Meta-DM can achieve significant improvements over various algorithms, including some that even outperform state-of-the-art methods, particularly on smaller datasets such as \textit{mini}ImageNet. We also perform ablation studies to make the experiments more plausible. We believe that this straightforward yet powerful module can inspire new approaches to tackle FSL problems.

\paragraph{Limitations and Future Work} We acknowledge that the effectiveness of Meta-DM has only been demonstrated on FSL problems in this paper. Although Meta-DM has shown great potential in improving the performance of FSL, it remains to be seen whether it can be applied to other image classification challenges. In the future, we plan to investigate the potential of Meta-DM in other areas such as fine-grained image recognition. We believe that our findings will contribute to the development of image recognition and other related fields.

\end{document}